\documentclass[12pt]{article}

\usepackage[english]{babel}
\usepackage[utf8]{inputenc}
\usepackage{csquotes}
\usepackage{graphicx}
\usepackage{caption}
\usepackage{subcaption}
\usepackage{dsfont}
\usepackage{verbatim}
\usepackage{authblk}
\usepackage{breakcites}
  
\usepackage[tbtags]{amsmath}
\usepackage{amsfonts}
\usepackage{dsfont}
\usepackage{bm}
\usepackage{url}
  
\usepackage{mathtools}
\usepackage{a4wide}

\usepackage[linesnumbered]{algorithm2e}
\usepackage{algorithmic}
\usepackage{soul}
\usepackage{placeins}
\usepackage{float}

\usepackage{array}
\usepackage{makecell}
\usepackage[labelfont=bf]{caption}

\usepackage[symbol]{footmisc}

\usepackage{xr}

\makeatletter
\DeclareFontEncoding{LS1}{}{}
\DeclareFontSubstitution{LS1}{stix}{m}{n}
\DeclareMathAlphabet{\mathscr}{LS1}{stixscr}{m}{n}
\makeatother

\title{Optimized spiking neurons classify images with high accuracy through temporal coding with two spikes
\\
}

\author
{Christoph Stöckl$^{1}$, Wolfgang Maass$^{1,\ast}$}

\begin{document}

\maketitle
\newenvironment{affiliations}{%
    \setcounter{enumi}{1}%
    \setlength{\parindent}{0in}%
    \slshape\sloppy%
    \begin{list}{\upshape$^{\arabic{enumi}}$}{%
        \usecounter{enumi}%
        \setlength{\leftmargin}{0in}%
        \setlength{\topsep}{0in}%
        \setlength{\labelsep}{0in}%
        \setlength{\labelwidth}{0in}%
        \setlength{\listparindent}{0in}%
        \setlength{\itemsep}{0ex}%
        \setlength{\parsep}{0in}%
        }
    }{\end{list}\par\vspace{12pt}}

\begin{affiliations}
\item {Institute of Theoretical Computer Science, Graz University of Technology, \\
Inffeldgasse 16b, Graz, Austria \\
$^\ast$ To whom correspondence should be addressed; E-mail: maass@igi.tugraz.at.}
\end{affiliations}
\begin{abstract}
Spike-based neuromorphic hardware promises to reduce the energy consumption of image classification and other deep learning applications, particularly on mobile phones or other edge devices. However, direct training of deep spiking neural networks is difficult, and previous methods for converting trained artificial neural networks to spiking neurons were inefficient because the neurons had to emit too many spikes. We show that a substantially more efficient conversion arises when one optimizes the spiking neuron model for that purpose, so that it not only matters for information transmission how many spikes a neuron emits, but also when it emits those spikes. This advances the accuracy that can be achieved for image classification with spiking neurons, and the resulting networks need on average just two spikes per neuron for classifying an image.
In addition, our new conversion method improves latency and throughput of the resulting spiking networks.
\end{abstract}

Spiking neural networks (SNNs) are 
currently explored as possible solution for a major impediment of more widespread uses of modern AI in edge devices: The energy consumption of the large state-of-the-art artificial neural networks (ANNs) that are produced by deep learning. 

This holds in particular for the Convolutional Neural Networks (CNNs) that are commonly used for image classification, but also other application domains.
These ANNs have to be large for achieving top performance, since they need to have a sufficiently large number of parameters in order to absorb enough information from the huge data sets on which they have been trained, such as the 1.2 million images of the ImageNet2012 dataset.
Inference with standard hardware implementations of these large ANNs is inherently power-hungry \cite{Garcia-Martin2019}. 

\begin{figure}[H]
\centering
 \includegraphics[scale=1.0]{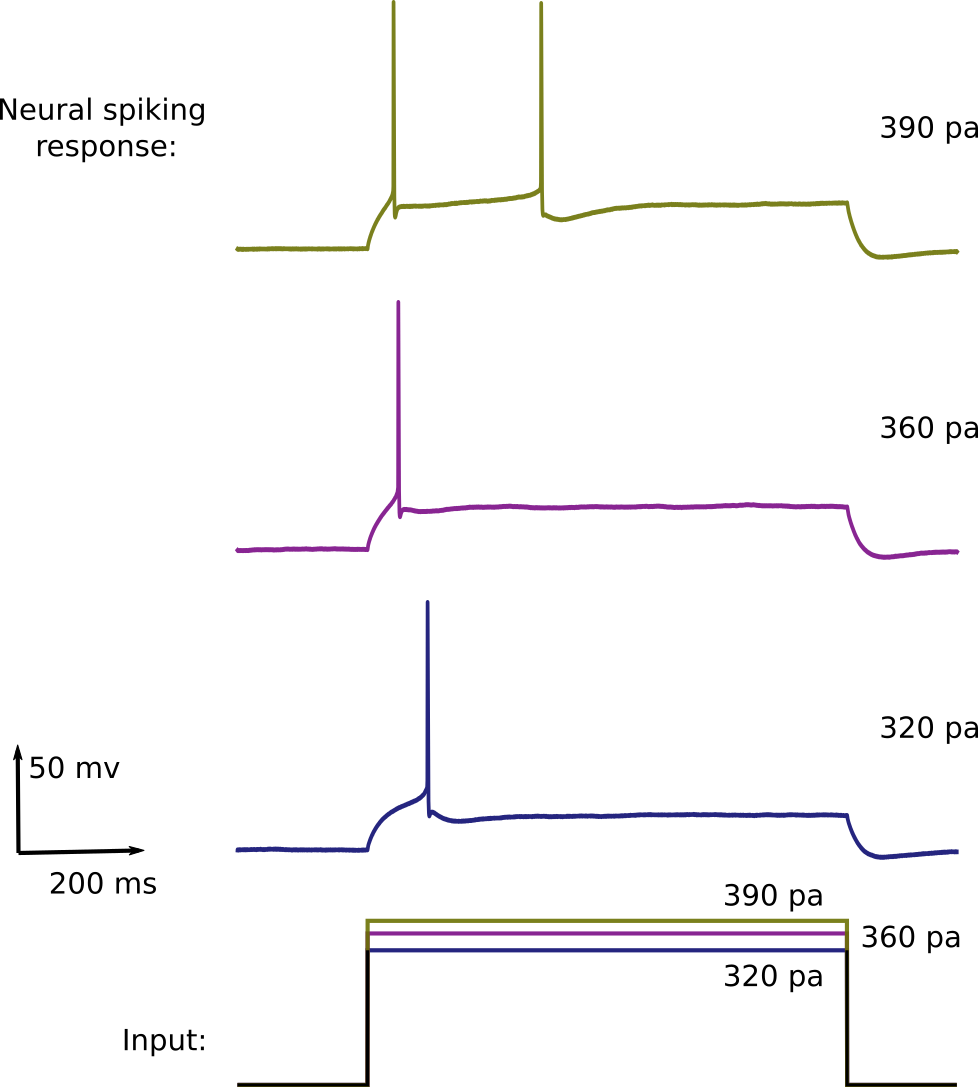}

\caption{\textbf {Encoding of different input values (current steps of different amplitudes) by temporal spike patterns in  a biological neuron.} 
\textit{Data taken from the Allen Cell Type Database$^1$ (Layer $3$ spiny neuron from the human middle temporal gyrus). }}
\label{bio-inspiration}
\end{figure}

Spiking neurons have been in the focus of the development of novel computing hardware for AI with a 
drastically reduced energy budget, partially because the giant SNN of the brain –consisting of about 100 billion 
neurons-- consumes just 20W \cite{LingJ2001}. 
Spiking neurons output trains of stereotypical 
pulses  that are called spikes. Hence their output is very different from the continuous numbers that an ANN neuron produces as output. Most spiking neuron models that are considered for implementation in neuromorphic
hardware are 
inspired by simple models for spiking neurons in the brain. 
However, these simple neuron models do not capture the capability of biological neurons to 
encode different inputs by different temporal spike patterns, not just by their firing rate 
(see Fig.~\ref{bio-inspiration} for an example). 

\footnotetext{$1$\ \textcopyright\ 2015 Allen Institute for Brain Science. Allen Cell Types Database. Available from: https://celltypes.brain-map.org/experiment/electrophysiology/587770251}

While large ANNs, trained with ever more sophisticated deep learning algorithms on 
giant data sets, approach --and sometimes exceed-- human performance in several categories of intelligence,
the performance of the current generation of spike-based neuromorphic hardware is lagging behind. 
There is some hope that this gap can be closed for the case of recurrent spiking neural networks, since 
those can be trained directly to achieve 
most of the performance of 
recurrent ANNs 
\cite{Bellec2020}. 

But the problem to produce SNNs that achieve similar performance as ANNs with few spikes persists for feedforward networks. Feedforward CNNs that achieve really good image classification 
accuracy tend to be very deep and very large, and training corresponding deep and large feedforward SNNs has not been 
able to 
reach similar classification accuracy.
Problems with the timing of spikes and precision of 
firing rates on higher 
levels of the resulting SNNs have been cited as possible reasons. 
One attractive alternative 
is to simply take a well-performing trained CNN 
and convert it into an SNN –using the same connections and weights. 
The most common –and so far best performing— conversion method was based on the idea of (firing-) rate coding, where the analog
output of an ANN unit is emulated by the firing rate of a spiking neuron \cite{Rueckauer2017}.
This method 
had produced so far the best SNN results for image classification. 
But the transmission of an analog value through a firing rate tends to require 
a fairly large number of spikes,  which reduces both latency and throughput of the network. 
Furthermore, the resulting SNN tends to produce so many spikes that its energy-advantage over non-spiking hardware 
gets lost. Finally, a rate-based ANN-to-SNN conversion can not be applied to those ANNs that currently achieve the highest accuracy on ImageNet, EfficientNets \cite{Tan2019}, because these employ an activation function that assumes both positive and negative values: the SiLU function 
\cite{hendrycks2016gaussian}.

We introduce a new ANN-to-SNN conversion 
that we call FS-conversion because it requires a spiking neuron to emit just a few spikes (FS = Few Spikes).  
This method is completely different from rate-based conversions,
and exploits the option of 
temporal coding with spike patterns, where the timing of a spike transmits extra information. 

Most previously proposed forms of temporal coding, see e.g. \cite{maass1998}, \cite{Thorpe2001}, \cite{Rueckauer2017}, \cite{kheradpisheh2020s4nn}, have turned out to be difficult to implement efficiently in 
neuromorphic hardware because they require to transmit fine time-differences between spikes 
to downstream neurons. 
In contrast, an FS-conversion 
can be implemented with just $log\ N$ different values of spike times and at most $log\ N$ spikes for transmitting integers between $1$ and $N$. 
Practically, the required number of spikes can be made even lower because not all N values occur equally often.
However FS-conversion requires a modified spiking neuron model, the FS-neuron, 
which has an internal dynamic that is optimized for emulating particular types of ANN neurons with few spikes.
We demonstrate the performance of SNNs that result from FS-conversion of CNNs, on two state-of-the-art datasets for image classification: ImageNet2012 and CIFAR10. This optimized spiking neuron model could serve as guidance for the next generation of neuromorphic hardware.

\section*{Emulating an ANN neuron by a spiking neuron with few spikes}
\begin{figure}[H]
\centering
 \includegraphics[scale=1.0]{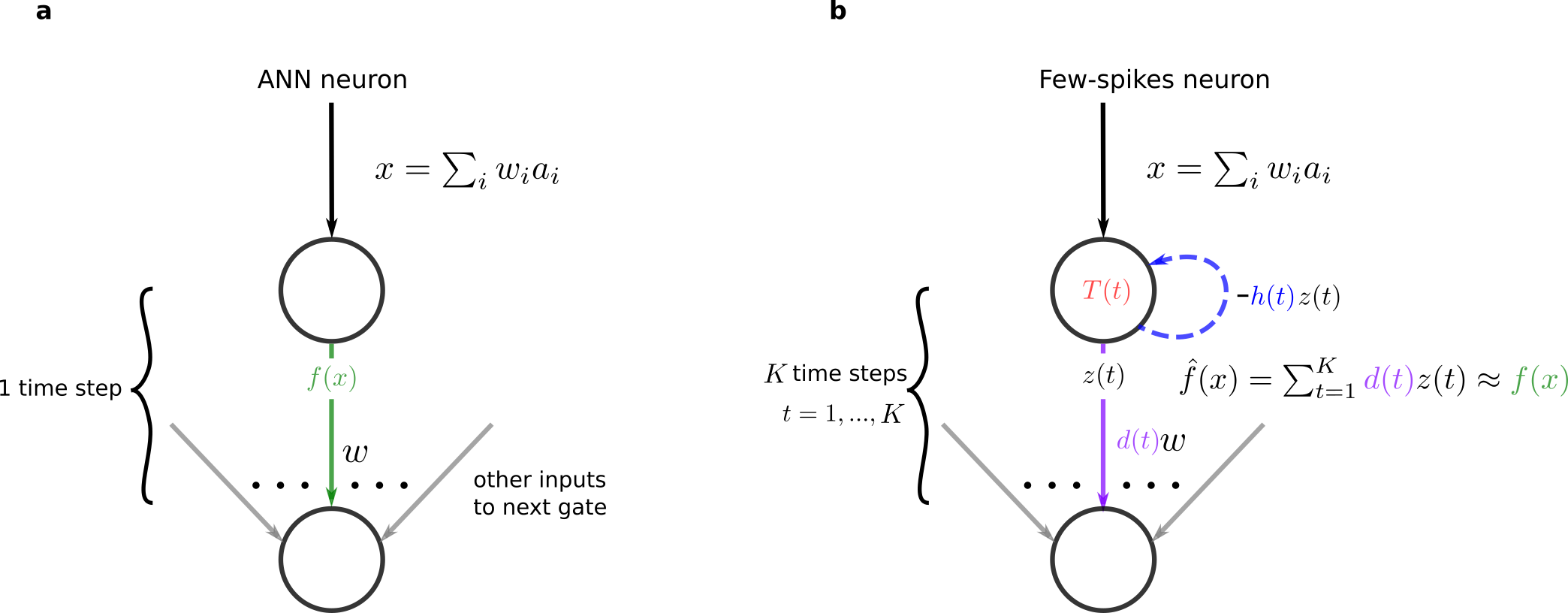}
\caption{\textbf{Conversion of an ANN neuron into an FS-neuron.} \\
\textit{
\textbf{a)} A generic ANN neuron with activation function $f(x)$ that is to be emulated. 
\\
\textbf{b)} An FS-neuron which emulates 
this ANN neuron in $K$ time steps $t = 1,...,K$. Its output spike train is denoted by $z(t)$.}}
\label{fs-coding}
\end{figure}

The FS-conversion from ANNs to SNNs requires a variation of the standard spiking neuron model, 
to which we refer as FS-neuron. 
The computation step of a generic artificial neuron in an ANN (see Fig. \ref{fs-coding} a) is emulated by $K$ time steps of an FS-neuron (Fig. \ref{fs-coding} b). Its internal dynamics is defined by fixed parameters $T(t), h(t), d(t)$ for $t = 1, ...,K$. These are optimized to emulate the activation function $f(x)$ of the given ANN neuron by a weighted sum of spikes $\sum_{t = 1}^K d(t)z(t)$, where $z(t)$ denotes the spike train that this neuron produces. More precisely: $z(t) = 1$ if the neuron fires at step $t$, else $z(t) = 0$.  
To emit a spike at time $t$, a neuron's membrane potential $v(t)$ has to surpass the current value $T(t)$ of its firing threshold.
We assume that the membrane potential $v(t)$ has no leak, but  is  reset 
to $v(t) - h(t)$ after a spike at time $t$. 
Expressed in formulas, the membrane potential $v(t)$ starts with value $v(1) = x$ where $x$ is the gate input, and  evolves during the $K$ steps according to

\begin{equation}
v(t+1) = v(t) - h(t)z(t).                      
\end{equation}
The spike output $z(t)$ of an FS-neuron for gate input $x$ can be defined compactly by 
\begin{equation}
 z(t) = \Theta (v(t) - T(t)) = \Theta \left( \left( x - \sum_{j=1}^{t-1} h(j)z(j) \right ) - T(t) \right ), \quad  t = 1,...,K ,
\end{equation}
where $\Theta$ denotes the Heaviside step function. 
The total output $\hat f(x)$ of the FS-neuron from these $K$ time steps, which is collected by the FS-neurons on the next layer, can be written as:
\begin{equation}
\hat f(x) =  \sum_{t = 1}^ K  d(t) z(t). 
\end{equation}
An illustration of the 
model can be found in Fig. \ref{fs-coding}b.

\begin{figure}[H]
\centering
 \includegraphics[scale=1.0]{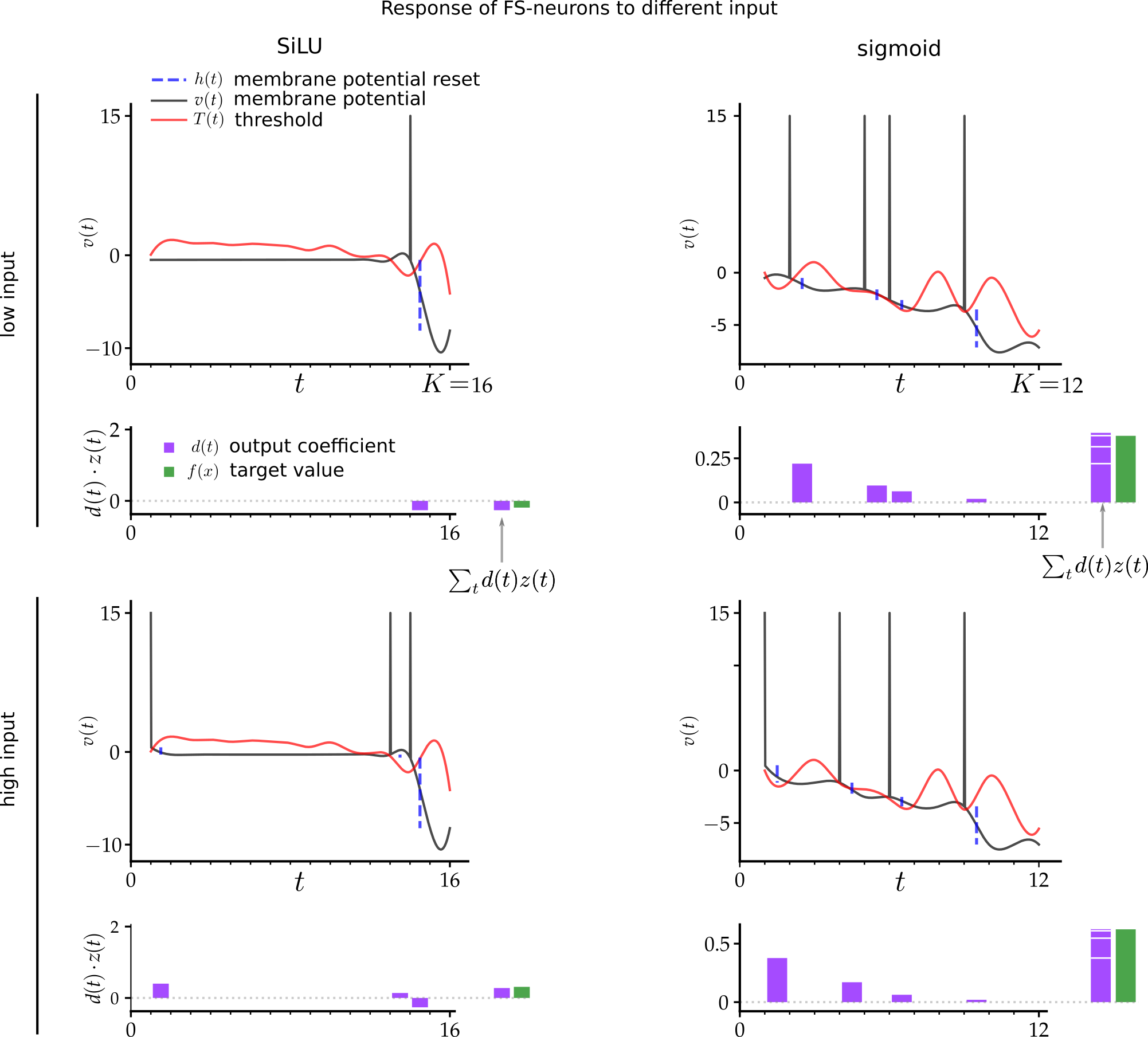}\\
\caption{\textbf{Internal dynamics of Few-Spikes neurons.}\\
\textit{The first row depicts the response of the FS-neurons 
to a low input value ($x=-0.5$) and the second row displays the response to a high input ($x=0.5$). The first column shows responses of an SiLU FS-neuron, while the second column a sigmoid FS-neuron. 
The relevant values of $T(t)$ and $v(t)$ for discrete time steps $t$ (see Fig. \ref{fs-swish}b and d) are 
smoothly interpolated for illustration.}} 
\label{fig:fs-bio-all}
\end{figure}

For emulating the ReLU activation function one can choose the parameters of the FS-neuron so that they define a coarse-to-fine processing strategy for all input values $x$ that lie below some upper bound, as described in the Methods section.  For emulating the SiLU function of EfficientNet one achieves a better FS-conversion if the parameters are chosen in such a way that they enable iterative –and thereby more precise-- processing for the range of inputs between $-2$ and $2$ that occur most often as gate inputs $x$  in EfficientNet. The resulting dynamics of FS-neurons is illustrated in Fig.~\ref{fig:fs-bio-all} for the case of the SiLU and sigmoid activation functions.

All FS-neurons that emulate ANN neurons with the same activation function can use the same parameters $T(t)$, $h(t)$, $d(t$), while the factor $w$ in the weights of their output spikes is simply lifted from the corresponding synaptic connection in the trained ANN (see Fig. \ref{fs-coding}).

Note that the number of neurons and connections in the network is not increased through the FS-conversion. However the number of computation steps $L$ of a feedforward ANN with $L$ layers is increased by the factor $K$. But the computations of the ANN can be emulated in a pipelined manner, where a new network input (image) is processed every 2$K$ time steps by the SNN. In this case the parameters of the FS-neurons change periodically with a period of length $K$ while the FS-neurons compute. These $K$ steps are followed by $K$ time steps during which the FS-neurons are inactive,  while the FS-neurons on the next layer collect their spike inputs for emulating the next computation step or layer of the ANN. Note that since all FS-neurons that emulate ANN neurons with the same activation function can use the same parameters $T(t)$, $h(t)$, $d(t)$, they require only little extra memory on a neuromorphic chip. \\
Both the TensorFlow code and the chosen parameters of the FS-neurons are available online\footnote{https://github.com/christophstoeckl/FS-neurons}.

\section*{Application to ImageNet}
The ImageNet data set \cite{Russakovsky2015} has become the most popular benchmark
for state-of-the-art image classification in machine learning (we are using here the ImageNet2012 version).
This data set consists of $1.281.167$ training images and $50.000$ test images 
(both RGB images of different sizes), that are labeled by $1000$ different categories. 
Classifying images from ImageNet is a nontrivial task even for a human, since this data
set contains for example $59$ categories for birds of different species and gender \cite{van2015building}. 
This may explain why a relaxed performance measurement, where one records whether the 
target class is among the top $5$ classifications that are proposed by the neural network ("Top5"),
is typically much higher.

\begin{figure}[H]
\centering
 \includegraphics[scale=1.0]{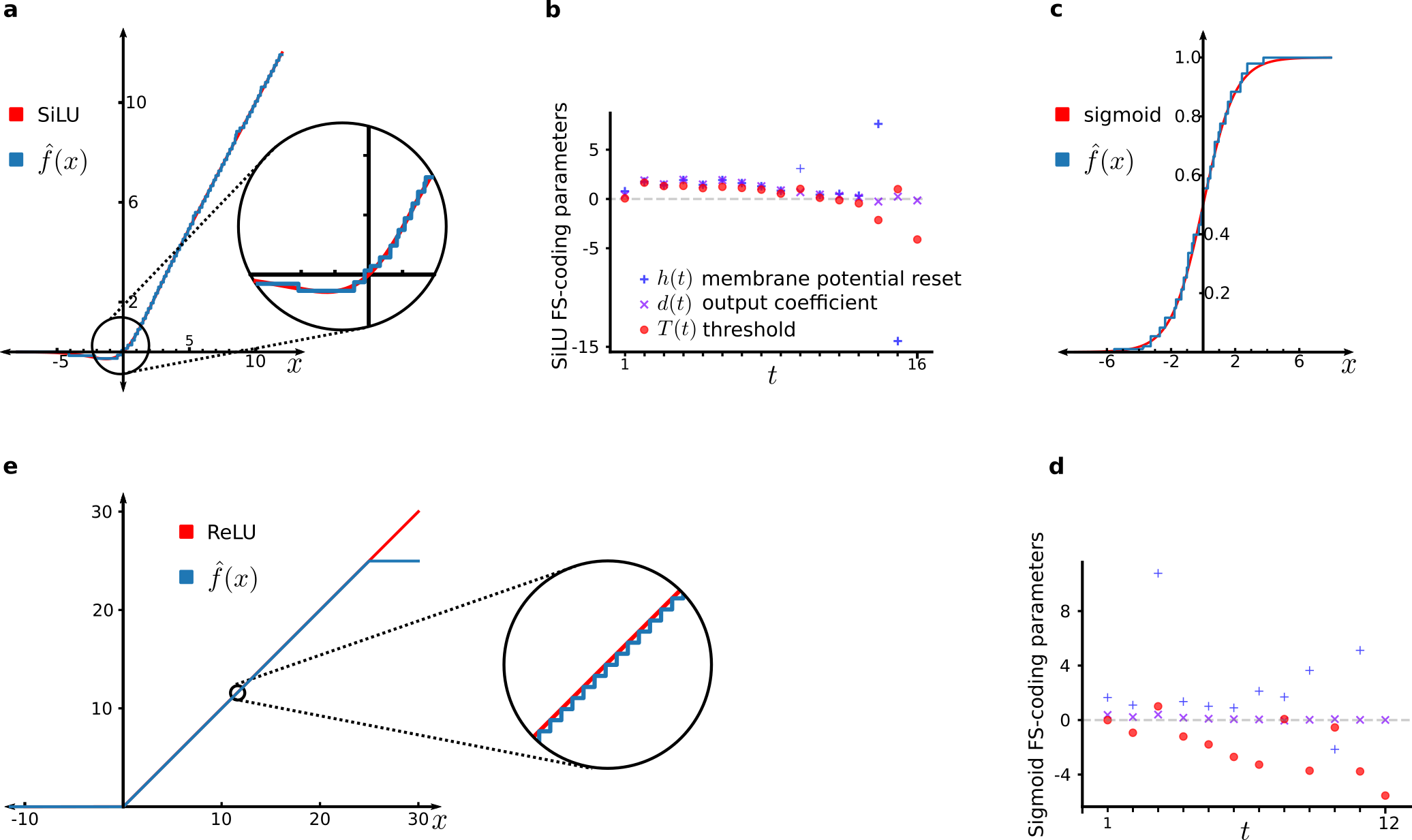}\\
\caption{\textbf{Approximations $\bm{\hat f(x)}$ of different activation functions by FS-neurons} \\
\textbf{a)} \textit{Approximation of the SiLU function with a single FS-neuron.\\
(red: SiLU function, blue: FS-approximation with $K=16$)}\\
\textbf{b)} \textit{Optimized internal parameters of the SiLU FS-neuron.} \\
\textbf{c)} \textit{Approximation  of the sigmoid function with a single FS-neuron.} \\
\textbf{d)} \textit{Optimized internal parameters of the sigmoid FS-neuron.} \\
\textbf{e)}  \textit{Approximation of the ReLU function with $K=10$ and $\alpha=25$.}}
\label{fs-swish}
\end{figure}

The recently proposed EfficientNet \cite{Tan2019} promises to become a new standard CNN architecture 
due to its 
very high accuracy while utilizing 
a smaller number of parameters than other CNN architectures.
EfficientNet uses as activation function $f(x)$ besides the SiLU function (Fig. \ref{fs-swish}) also the 
familiar sigmoid 
function, 
shown as the red curve in Fig. \ref{fs-swish} c. 
Note that $99.97\%$ of its activation functions are SiLU functions, 
making the appearance of the sigmoid function comparatively rare. 
The SiLU function emerged from preceding work on optimizing activation functions in ANNs \cite{Zoph2018}.
Another characteristic of the EfficientNet architecture is the extensive usage of depth-wise 
separated convolution layers. 
In between 
them, linear activation functions are used. Although it would certainly be possible to approximate linear functions using FS-coding, we
simply collapsed linear layers into the generation of the weighted sums that form the inputs to
the next layers.

Since the SiLU function assumes also negative values, it appears to be difficult to convert an ANN with this activation function via rate-coding to a spiking neuron. But it is fairly easy to convert it to an FS-neuron.
The values of the parameters 
$T(t), h(t) \text{ and } d(t)$ for $t = 1, ..., K$ of the FS-neuron can 
be obtained by training the FS-neuron model to fit the SiLU function, see Fig. \ref{fs-swish} a and b. 
We used for that backpropagation through time, with 
a triangle-shaped pseudo derivative for the 
non-existing derivative of the Heaviside step function. 

In most cases, the possible inputs to an activation function are not uniformly distributed, 
but there exists a certain region in which most inputs lie with high probability.
For example, most of the inputs to the SiLU functions in the EfficientNet are
in the interval from $-2$ to $2$ and therefore, achieving a high approximation accuracy in this region is especially desirable, (see Fig. \ref{spikes-and-performance} a).
It is possible to encourage the FS-neuron to put more emphasis on a certain region, by assigning a high weight in the loss function to this region. More details about the training procedure of the FS-parameters can be found in the Methods section.

The effective activation function of the resulting FS-neuron is shown in Fig. \ref{fs-swish}a. Fig. \ref{fs-swish} c shows the corresponding result for the FS-conversion of an ANN neuron with the sigmoid activation function.

\begin{table}[H]
\centering
\def\arraystretch{1.5}
\begin{tabular}{|c|c|c|c|c|c|c|}
\hline
Model &  \makecell{ANN \\accuracy} &\makecell{accuracy of the\\SNN produced \\by FS-conversion}& \# params & \# layers & \# neurons & \# spikes \\ \hline
\multicolumn{7}{|c|}{ImageNet2012} \\ \hline
EfficientNet-B7 &  \makecell{$85$\% \\ (97.2 \%)} & \makecell{83.57\% \\ (96.7\%)} &$66$M & 218 & 259M & 554.9M \\ \hline
ResNet50 & \makecell{75.22\% \\ (92.4\%)} & \makecell{75.10\% \\ (92.36\%)} & 26M & 50 & 9.6M & 14.045M \\ \hline
\multicolumn{7}{|c|}{CIFAR10} \\ \hline
ResNet8 & 87.22\% & 87.05\% &78k & 8 &73k & 103k \\ \hline
ResNet14 & 90.49\% & 90.39\% &174k & 14 &131k & 190k \\ \hline
ResNet20 & 91.58\% & 91.45\% &271k &  20 &188k & 261k \\ \hline
ResNet50 & 92.99\% & 92.42\% &755k & 50 & 475k & 647k \\ \hline
\end{tabular}
\caption{\textbf{Accuracy and spike numbers for classifying images from ImageNet with FS-conversions of two state-of-the-art CNNs.}\textit{The SNNs produced by FS-conversion of the ANNs achieved almost the same accuracy, and usually used at most $2$ spikes per neuron. Top5 accuracy is reported in parentheses. 
The number of spikes needed for inference was obtained by averaging over the 1000 test images.}}

\label{tab:imgnetres}
\end{table}

Using these FS-neurons it is possible to 
emulate the EfficientNet-B7 model with spiking neurons. 
The
accuracy of the resulting spiking CNN, using the publicly available weights $w$ of the trained EfficientNet, can be found in Table \ref{tab:imgnetres},
together with the total number of spikes that it uses for sample inferences. 

The FS-conversion of EfficientNet-B7 achieved an accuracy of $83.57\%$. The best accuracy for ImageNet that had previously been reported for SNNs was $74.6\%$ \cite{Rueckauer2017}. It was achieved by a rate-based conversion, which required a substantial number of spikes per neuron and about 550 time steps for each image classification. The SNN resulting from FS-conversion of 
EfficientNet-B7 used about $2$ spikes per neuron for classifying an image.
The FS-neurons approximating the SiLU function
used $K=16$ and the FS-neurons approximating the sigmoid function used $K=12$.

The 
layers of the CNN that use the SiLU function as activation function can be simulated in a pipelined manner by the SNN, processing a new image every $2K$ time steps: Its first $K$ time steps are spent collecting the outputs from the preceding layer of FS-neurons during their $K$ time steps of activity. It then processes these collected inputs $x$ during the subsequent $K$ time steps.

Hence the SNN that results from FS-conversion of EfficientNet can classify a new image every $2K = 32$ time steps.
Further implementation details can be found in the Methods section.

\subsection*{Approximating the ReLU activation function}
The ReLU activation function, see Fig.~\ref{fs-swish} d, is among the most frequently used activation functions,
and also quite good accuracies have been achieved with it for ImageNet.
It represents a special case for FS-conversion, as it is possible to 
find the ideal values for $h(t), T(t) $ and $ d(t)$ analytically, bases on the idea of computation with binary numbers. 
By setting the parameters of the FS-neuron to 
$T(t) = h(t) = d(t) = 2^{K-t}$, the FS-neuron approximates the ReLU activation function $f(x)$ with a coarse-to-fire-processing strategy.
Let us assume for simplicity that an FS-neuron receives inputs $x$ from $(-\infty, 0] \cup \{1, 2, ..., 2^K-1\}$.  
Then it reproduces with the specified parameters the output ReLU($x$) of the ReLU gate for any $x$ from $(-\infty, 0] \cup \{1, 2, ..., 2^K-1\}$ without error. 
In order to be able to transmit also non-integer values $x$ between $0$ and some arbitrary positive constant $\alpha$, one simply multiplies the given values for $T(t), h(t) \text{ and } d(t)$ with $\alpha 2^{-K}$. 
Then the FS-neuron reproduces ReLU($x$) for any non-negative $x$ less than $\alpha$ that are multiples of $\alpha 2^{-K }$ without error, and ReLU($x$) is rounded down for values $x$ in between to the next larger multiple of $\alpha 2^{-K}$.
Thus the output of the FS-neuron deviates for $x$ in the range from $-\infty$ to $\alpha$ by at most $\alpha 2^{-K}$ from the output of the ReLU gate.
The resulting approximation is plotted for $\alpha=10$ in Fig. \ref{fs-swish} d.
Several advantages arising from the simple structure of the parameters have been laid out in the Methods section.

The accuracy of 75.22\%  for the ANN version of ResNet50 in Table \ref{tab:imgnetres} resulted from 
training a variant of ResNet50 where max-pooling was replaced by average pooling, using
the hyperparameters given in the TensorFlow repository\footnote{https://github.com/tensorflow/tpu/tree/master/models/official/efficientnet}. 
The resulting accuracy in ImageNet is close to the best published performance of 76\% for ResNet50 ANNs \cite[Table 2]{Tan2019}. 
The application of the FS-conversion to this variant of ResNet50 (with $K = 10$ and $\alpha = 25$) yields an SNN whose Top1 and Top5 performance is almost indistinguishable from that of the ANN version. 

\section*{Application to CIFAR10}
CIFAR10 \cite{krizhevsky2009learning} is a smaller and more frequently used dataset for image classification. It consists of 60.000 colored images, each having a resolution of just 32 by 32 pixels, and just 10 image classes. 
The results for ANN versions of ResNet that are given in Table \ref{tab:imgnetres} for CIFAR10 
arise from training them with the hyperparameters given in the TensorFlow models repository.
They use the ReLU function as the only nonlinearity, since we have replaced there max-pooling by average pooling. Nevertheless, they achieve an accuracy for CIFAR10 which is 
very close to the best results reported for CIFAR10 in the literature.
The best performing reported ResNet on CIFAR10 is ResNet110, where a test accuracy of 93.57\% had been achieved \cite{He2016}.
Our ResNet50 achieves 92.99\%, which is similar to their accuracy 
of 93.03\% for ResNet56.

\section*{Analysis of FS-coding}
\begin{figure}[H]
\centering
 \includegraphics[scale=1.0]{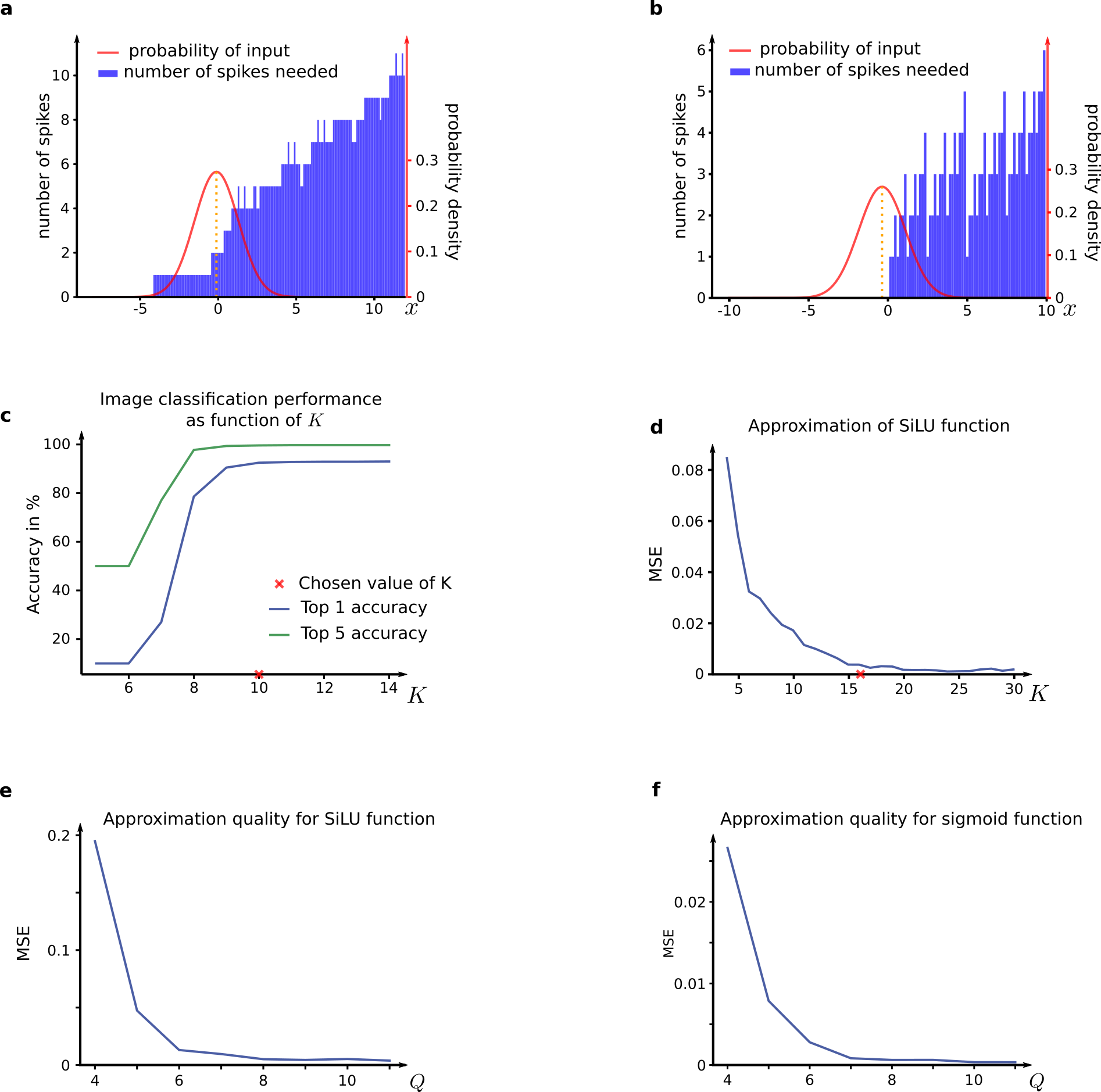}
\caption{\textbf{Number of spikes needed by FS-neurons for image classification and influence of $K$ and bit precision $Q$ on performance}\\
\footnotesize
\textbf{a)} \textit{The number of spikes used by an a FS-neuron with $K = 16$ to approximate the SiLU function, as function of its input value $x$. 
The red Gaussian 
models the probability that the FS-neuron will receive this input value in the EfficientNet-B7 model (mean = $-0.112$, variance = $1.99$).}
\textbf{b)}\textit{ The number of spikes 
used by an FS-neuron to approximate the ReLU function 
with $K=6$ and $\alpha=10$.
The red Gaussian 
models the probability that the FS-neuron will receive this input value in the 
ResNet50 model in an application to images from ImageNet (mean $-0.36970$, variance = $2.19$).}
\textbf{c)} \textit{Test Accuracy of the ResNet50 model on CIFAR10 with 
FS-neurons, in dependence on $K$. The red cross indicates the chosen value of $K$ for our results.} 
\textbf{d)} \textit{Mean squared error (MSE) of a SiLU approximation by FS-neurons with different values of $K$. 
The red cross indicates the chosen value of $K$ in the given context.} 
\textbf{e)} \textit{MSE of a SiLU approximation by FS-neurons with $K=16$ as function of the bit precision $Q$ of its parameters.} 
\textbf{f)} \textit{MSE of a sigmoid approximation by FS-neurons with $K=12$ 
as function of the bit precision $Q$ of its parameters. \normalsize}}
\label{spikes-and-performance}
\end{figure}

On digital neuromorphic hardware the energy consumption is proportional to the number 
of spikes which are needed for a computation. 
The number of spikes needed for an FS-neuron to perform the approximation of the 
target function is 
depicted in Fig. \ref{spikes-and-performance} a and b as function of the gate input $x$.
If one compares these numbers with the distribution of input values $x$ (red curves) that typically occur during image classification, one sees why on average less than $2$ spikes are used by FS-neurons for these applications.

The most important specification of an FS-neuron is the number $K$ of time steps that it uses. 
Fig. \ref{spikes-and-performance}c, d provide insight into the nature of the trade-off between the size of $K$ and the approximation quality of the FS-neuron. 

Furthermore, it is of interest to consider scenarios where only a certain 
number of bits are available for the FS-neuron parameters.
To analyze the impact of 
that we consider a setting where 
the parameters of the FS-neurons can only take on discrete 
values in the range from $[-8, 8]$. 
The possible values are equally spaced and the 
number of values can be written as $2^Q$, where $Q$ refers to the number 
of bits which are available for each parameter $T(t)$, $h(t)$, $d(t)$ of the FS-neuron. 
Fig. \ref{spikes-and-performance}e, f depict the impact of such quantization 
on the mean squared error of the approximation of the activation function.

\subsection*{Expected implementation cost on neuromorphic hardware}
We distinguish three types of neuromorphic hardware

\begin{itemize}
	\item Digital, but hardware not constrained to a particular neuron model (example: SpiNNaker)
	\item Digital, but hardware is constrained to a particular neuron model (example: Loihi)
	\item Mixed digital analog (examples: IBM research chip with memristors and BrainScaleS-2)
\end{itemize}

\subsubsection*{SpiNNaker}
The SpiNNaker platform \cite{furber2014spinnaker} provides a flexible environment which is not constrained to a specific neuron model. 
SpiNNaker allows to compute all products $d(t)w$ on the chip, which reduces the additional memory consumption to to a small constant value.
All parameters $T(t)$, $h(t)$ and $d(t)$ only need to be stored in memory once, as they can be shared across all neurons which approximate the same activation function.
The additional computational complexity of the FS-neuron model also has a very small impact, as computing the updated weight $d(t)w$ can be done with a single instruction cycle.

\subsubsection*{Loihi}
Loihi \cite{davies2018loihi} also promises to be an interesting target platform for FS-neurons. Especially FS-neurons approximating the ReLU activation function could be ported very efficiently
to this hardware platform. As the chip is based on fixed-point arithmetic, 
one can implement $T(t)$, $h(t)$ and $d(t)$ for ReLu using
a single parameter, namely the shared weight exponent. This is be possible due to the fact
that at every time step $t$ all FS-parameters have the same value, which is always a power of $2$. Therefore, the additional memory consumption does not grow with $K$. It is also possible to use other activation functions besides ReLU on Loihi, however, in this case it would be necessary to store all products $d(t)w$ on the chip, as computing the updated weight in an online fashion is not possible.
In this case, an increase in memory consumption of $\mathcal{O}(K)$ is expected.

\subsubsection*{IBM research chip with memristors}
IBM has presented an in-memory chip architecture supporting both ANNs and SNNs in the Supplementary Material S3 of the article \cite{wozniak2020deep}. This architecture employs a memristor crossbar array for fast (time complexity $\mathcal{O}(1)$) and energy-efficient multiplication of the outputs of one layer $l$ of neurons with the weights of synaptic connections to neurons on the next layer $l+1$. One can replace all spikes (i.e., values $1$) that emerge from layer $l$ at time $t$ of the $K$-step cycle in the emulation of the neurons on layer $l$ by a common value value $d(t)$ that  is centrally stored.  Since  the values $d(t)$ and $0$ can be used directly as inputs to the memristor array, no significant extra cost is expected. The neuron models are implemented in the digital part of this neuromorphic chip architecture of IBM, using very fast digital logic and SRAM for storing parameters. Since all neurons on a layer $l$ of our FS networks use the same parameters $T(t)$ and $h(t)$, they can be stored in a local SRAM for all neurons on layer $l$, in a similar fashion as on SpiNNaker.
A neuron model that goes already one step in the direction from LIF to FS-neurons has actually already been implemented on this architecture: The soft spiking neural unit (sSNU), that emits analog instead of binary values and subtracts a corresponding value from the membrane potential \cite{wozniak2020deep}. \\

\subsubsection*{BrainScaleS-2} 
This neuromorphic chip \cite{billaudelle2020versatile} is also a mixed analog digital architecture where a digital plasticity processor allows fast changes of synaptic weights, but also central memory storage and application of the time-varying parameters $T(t)$, $h(t)$ and $d(t)$ of the neuron dynamics. 
Like on SpiNNaker, the parameters only have to be stored once in memory and can be shared across many neurons.
The leak term of the membrane voltage of the analog neuron models can be switched off, so that the analog part can be used for efficient matrix multiplication in a similar manner as on the IBM chip.

\section*{Methods}
In this section various details necessary to reproduce our results have been listed. 
Additionally, to aid the interpretation of the results, a comparison to previous conversion approaches has been added. 

When training the parameters of the FS-neurons it is important to specify an interval in which 
the approximation should be very good. Ideally, most of the inputs to the ANN neuron should fall into this interval to 
guarantee a good performance.
In our experiments, the FS-neurons have been trained to approximate 
the interval from $[-8, 12]$ for the SiLU function and $[-10, 10]$ 
for the sigmoid function.
The resulting FS-neuron approximates the SiLU function with a
mean squared error of $0.0023$ inside the main region $[-2, 2]$ and $0.0064$ in 
the region outside, which can be written as $[-8, -2] \cup [2, 12]$.
As a result of our fine-tuning the values for $T(t)$, $d(t)$ and $h(t)$ stay 
for most time steps $t$ within the main region $[-2, 2]$ as can be seen in Fig. \ref{fs-swish}b. 

To reduce the complexity of the converted CNN, we decided not to emulate the multiplication operation by FS-neurons, 
which occurs in the CNN if the squeeze 
and excitation optimization \cite{Hu2018} is being used. 
In many neuromorphic chips, such as SpiNNaker and Loihi, the on-chip digital processor 
could carry out these multiplications. Otherwise one can approximate multiplication
in a similar manner as the SiLU function with 
a suitably optimized FS-neuron, see \cite{stoeckl2019}. Alternatively one can compute multiplication with a small circuit of
threshold gates, i.e., very simple types of spiking neurons, of depth 2 or 3. 
A recent summary of such results is provided in section 3 of \cite{parekh2018}.
\newline
\newline
Due to the simple structure of the parameters of the ReLU FS-neurons several advantages arise.
In particular when approximating the ReLU function with an FS-neuron it 
is possible to calculate the changes of parameters 
for $t = 1,...,K$ by simply using a bit shift operation,
possibly providing a very efficient implementation on neuromorphic hardware.
The resulting SNN can be used in a pipelined manner,
processing a new network input every $2K = 20$ time steps, analogously as for the SiLU function.

\subsection*{Further properties of FS-coding that are relevant for neuromorphic hardware}
One of the major advantages of using FS-neurons in neuromorphic hardware is the smaller amount 
of time steps and spikes required to approximate artificial neurons. 
For the case of the ReLU activation function, a rate coded spiking neuron requires 
$N$ time steps to encode $N$ different values. 
FS-neurons improve upon this unary coding scheme by utilizing the time dimension to implement a 
binary coding scheme. Therefore, the number of time steps required to encode $N$ different values can be reduced  to just $log_2(N)$. \\
To underline the binary coding nature of FS-neurons, in the case of the ReLU activation function, the corresponding FS-neurons will show a spiking pattern equivalent to of the output of the ReLU function, written as a binary number.
The same logarithmic relation holds not only for the number of time steps required but also for the number of spikes needed to transmit a value. More sophisticated codes could be used to make the computation robust to noise in spike transmission.

Note, that most of the inputs to the FS-neurons have a value close to $0$, as shown in Fig. \ref{spikes-and-performance} a and b. Consequently, the FS-neurons usually require only a few spikes to transmit the output values, making the the coding scheme even more sparse in practice. 

\subsection*{Comparison with previous methods}
The idea of converting a pre-trained ANN to a SNN has received a fair amount of attention in the 
recent years.
The most popular conversion approaches are rate-based, meaning they translate the continuous 
output of an artificial ReLU neuron into a firing rate of a spiking neuron. 
Unfortunately there are some drawbacks associated with rate-coding. 
Due to its unary coding nature, rate-codes are sub-optimal in the sense that they do not make 
good use of the time dimension. 
Usually a large amount of time steps is required to achieve a sufficiently accurate approximation. 
Furthermore, rate-based conversions are only capable of converting simple activation functions like ReLU, but fail to convert more sophisticated functions like SiLU.

Another popular conversion approach uses time to first spike (TTFS)  coding \cite{rueckauer2018conversion}. 
This approach encodes the continuous outputs of the corresponding ReLU ANN neurons in the length of the time interval until the first spike, resulting in a very sparse spiking activity. 
However, this method seems to not scale easily to large models and has, to the best of our knowledge, not been tested on large data sets like ImageNet. 
The idea of using single spike temporal coding has first been explored in \cite{maass1997fast} and it has been shown to have a variety of applications, like implementing an efficient $k$-NN algorithm on neuromorphic hardware \cite{frady2020neuromorphic}.

Furthermore, a new conversion method has been proposed, in which the spiking neurons can approximate 
the ReLU function using a hysteresis quantization method \cite{yousefzadeh2019conversion}. 
This approach waits to be tested on larger networks and datasets. 

One property that all previously mentioned conversion methods have in common is that 
they only consider transforming artificial ReLU neurons to spiking neurons, and therefore 
cannot convert more sophisticated activation functions, which are used in network architecture like the EfficientNets. 

A detailed summary comparing FS-coding to previous results can be found in the Extended Data Table $1$.

It is worth noting, that the throughput using FS-coding is substantially better than that of SNNs which result from rate-based ANN-to-SNN conversions of ANNs with the ReLU function, as proposed for example in \cite{Rueckauer2017, Sengupta2019}.
The Inception-v3 model in \cite{Rueckauer2017} 
was reported to yield a SNN that needed 550 time steps to classify an image.
Under the assumption that rate-based models profit only very little from pipelining, it is reasonable to estimate 
that the throughput of an SNN that results from FS-conversion of ReLU gates with $K=10$ is roughly $25$ times higher.

The SNN resulting from the rate-based conversion of the ResNet34 model discussed in \cite{Sengupta2019} 
has been reported to use $2500$ time steps for a classification.
Therefore we estimate that the throughput is increased here by a factor around $125$ through FS-conversion.

Spiking versions of ResNet20 have already been previously explored \cite{Sengupta2019}.
Using a rate-based conversion scheme an accuracy of 87.46\% was reported. 

FS-conversion of ResNet20 yields a substantially higher accuracy of 91.45\%, 
using just $80$ to $500$ time steps for each image -depending on the model depth- instead of $2000$, thereby significantly reducing latency. 
In addition, the throughput is drastically improved.

Also the number of spikes that the SNN uses for classifying an image from CIFAR10 is significantly reduced when one moves from a rate-based conversion to an FS conversion.
A converted ResNet11 has been reported to use more than $8$ million spikes to classify 
a single test example \cite{Lee2020}.
Comparing this to an FS-converted ResNet14 we find that the latter uses $40$ times fewer spikes despite being a slightly larger model. 
Using direct training of SNNs instead of a conversion scheme has been reported to 
result in a lower amount of spikes needed to perform a single classification.
However, even a directly trained SNN version of ResNet11 uses $7$ times more spikes than an 
FS-conversion of ResNet14 \cite[Table 8]{Lee2020}.

In \cite{rathi2020enabling} the authors present a novel approach for obtaining high performance 
SNNs by combining a rate-based conversion scheme with a subsequent gradient-based fine-tuning procedure. They report the highest accuracy for an SNN on CIFAR10, which was achieved by converting a very performant ANN. 
They also show results for ImageNet, where they achieve an accuracy of $65.1\%$ on their ResNet-$34$.
Deeper models, like the ResNet-$50$, were not considered in this work. 
On ImageNet FS-conversion of the ResNet-50 model improves their accuracy by 10\% and FS-conversion of the EfficientNet-B7 surpasses their performance by 18.47\%.

\section*{Discussion}
We have 
presented a new approach for generating SNNs that are very close to ANNs in terms of classification accuracy for images, while working in the energetically most attractive regime with very sparse firing activity. Besides substantially improved classification accuracy, they exhibit 
drastically improved latency and throughput compared with rate-based ANN-to-SNN conversions.
off the shelf.
One can argue that this is exactly the way which evolution has chosen for the design of neurons in living organism. Not only neurons with particular information processing tasks in the smaller nervous systems of insects, but also neurons in the neocortex of mammals exhibit an astounding diversity of genetically encoded response properties [ \cite{sterling2015principles},  
\cite{gouwens2019classification},
\cite{bakken2020evolution}]. In particular, the probability of producing a spike depends in diverse ways on the recent stimulation history of the neuron, see \cite{GerstnerETAL2014} for some standard models. In other words, the excitability of different types of biological neurons increases or decreases in complex ways in response to their previous firing. As a result, the temporal structure of a train of spikes that is produced by a biological neuron contains additional information about the neuron input that can not be captured by its firing rate. Similary, FS-neurons that are optimized for high accuracy image classification with few spikes exhibit history-dependent changes -encoded through their functions $T(t)$ and $h(t)$ according to equ.~(2)- in their propensity to fire, see Fig. ~\ref{fs-swish}b and \ref{fs-swish}e. Furthermore the function $d(t)$ enables subsequent neurons to decode their spikes in a timing-sensitive manner. In these regards an FS-conversion from ANNs to SNNs captures more of the functional capabilities of spiking neurons than previously considered rate-based conversions to an off-the-shelf spiking neuron model.

It is well known that spikes from the same neurons in the brain can transmit different information to downstream neurons depending on the timing of the spike, see e.g. phase precession in the hipppocampus \cite{harris2002spike}. Hence is is conceivable that downstream neurons give different weights to these spikes, in dependence of the firing time of the presynaptic neuron. In fact, it is well known that the large repertoire of pre- and postsynaptic synaptic dynamics found in different synapses of the brain \cite{markram2004interneurons, kopanitsa2018combinatorial} enables postsynaptic neurons to modulate the amplitude of postsynaptic responses in dependence of the timing of presynaptic spikes relative to an underlying rhythm. This can be viewed as a biological counterpart of the timing-dependent weights $d(t)$ in our model. Altogether we believe that FS-neurons provide a first step in exploring new uses of SNNs where information is not encoded by the timing of single spikes or firing rates, but by temporal spike patterns. 

Important for applications of FS-coding in neuromorphic hardware is that it is applicable to virtually any activation function, in particular to that activation function for ANN neurons that currently 
provides the highest accuracy on ImageNet, the SiLU function. Rate-based conversion can not be readily applied to 
the SiLU function because it assumes both positive and negative output values.
When approximating the more commonly used ReLU function, FS-neurons approach 
the information theoretic 
minimum of spikes for spike-based communication.

In fact, FS-neurons that emulate ANN gates with the ReLU activation function produce 
$1.5$ spikes 
on average for classifying an image, while those for the Switch activation function produce 2 spikes on average.
As the number of spikes required for inference by an SNN is directly related to its energy consumption in spike-based neuromorphic hardware, 
the energy consumption of FS-converted SNNs appears to be close to the theoretical optimum for SNNs. 
Since FS-conversion provides a tight bound on the number $K$ of time steps during which a spiking 
neuron is occupied, 
it can also be used for converting recurrently connected ANNs to SNNs. 

The proposed method for generating highly performant SNNs for image classification through FS-conversion of trained CNNs
offers an opportunity to combine the computationally more efficient and 
functionally more powerful training of ANNs with the superior energy-efficiency
of SNNs for inference.
Note that one can also use the 
resulting SNN as initialization for further 
training 
of the SNN, e.g., for a more specific task. 

Altogether our results suggest that spike-based hardware may gain an edge in the competition for the
development of drastically more energy-efficient hardware for AI 
if one does not forgot to optimize the spiking neuron model in the hardware for its intended range of applications. 
In contrast to energy efficient digital hardware that is optimized for specific ANN  architectures and activation functions, see e.g. \cite{zhang2019recent} for a review, a spike-based neuromorphic chip that is able to emulate FS-neurons can carry out inference for all possible ANN architectures.  It can also emulate ANNs with previously not considered activation functions, since a change of the activation function just requires reprogramming of the digitally stored parameters of FS-neurons. Hence such spike-based chip will be substantially more versatile than common digital hardware accelerators for AI.

\section*{Acknowledgements}
We would like to thank Franz Scherr for helpful discussions.
We thank Thomas Bohnstingl, Evangelos Eleftheriou, Steve Furber, Christian Pehle, Philipp Plank and Johannes Schemmell for advice regarding implementation aspects of FS-neurons in various types of neuromorphic hardware.
This research was partially supported by the Human Brain Project of the European Union (Grant agreement number 785907). 
We also thank our anonymous reviewers for very constructive comments. 
\section*{Competing Interests}
We are not aware of competing interests.

\section*{Data availability}
Both ImageNet \cite{deng2009imagenet} and CIFAR10 \cite{krizhevsky2009learning} are publicly available datasets. 
No additional datasets were generated or analysed during the current study.
The data for the spike response depiced in figure \ref{bio-inspiration} has been publishedbythe Allen Institute for Brain Science in 2015 (Allen Cell Types Database). Available from: https://celltypes.brain-map.org/experiment/electrophysiology/587770251

\section*{Author contributions}
CS conceived the main idea, CS and WM designed the model and planned the experiments, CS carried out the experiments, CS and WM wrote the paper. 

\section*{Code availability}
The code this work is based on is publicly available at: https://github.com/christophstoeckl/FS-neurons (DOI: 10.5281/zenodo.4326749).
Additionally, the code is also available in a Code Ocean compute capsule: https://codeocean.com/capsule/7743810/tree

\end{document}